# The Moral Machine Experiment on Large Language Models

**Kazuhiro Takemoto[1*]**

*1) Department of Bioscience and Bioinformatics, Kyushu Institute of Technology, Iizuka, Fukuoka 820-8502, Japan*
*\*Corresponding author's e-mail: takemoto@bio.kyutech.ac.jp*

## Abstract

As large language models (LLMs) become more deeply integrated into various sectors, understanding how they make moral judgments has become crucial, particularly in the realm of autonomous driving. This study utilized the Moral Machine framework to investigate the ethical decision-making tendencies of prominent LLMs, including GPT-3.5, GPT-4, PaLM 2, and Llama 2, comparing their responses to human preferences. While LLMs' and humans' preferences such as prioritizing humans over pets and favoring saving more lives are broadly aligned, PaLM 2 and Llama 2, especially, evidence distinct deviations. Additionally, despite the qualitative similarities between the LLM and human preferences, there are significant quantitative disparities, suggesting that LLMs might lean toward more uncompromising decisions, compared to the milder inclinations of humans. These insights elucidate the ethical frameworks of LLMs and their potential implications for autonomous driving.

# Introduction

Chatbots (e.g., ChatGPT [1], developed by OpenAI) are based on large language models (LLMs) and designed to understand and generate human-like text from the input they receive. As artificial intelligence (AI) technologies, including LLMs, become more deeply integrated into various sectors of society [2][3][4], their moral judgments are increasingly scrutinized. The influence of AI is pervasive, transforming traditional paradigms, and ushering in new ethical challenges. This widespread application underscores the importance of machine ethics, which mirrors human ethics [5]. Beyond the realm of traditional computer ethics, AI ethics probes further by examining the behavior of machines toward humans and other entities in various contexts [6].

Understanding AI's capacity for moral judgment is particularly crucial in the context of autonomous driving [7][8]. Since the automotive industry anticipates incorporating AI systems such as ChatGPT and other LLMs to assist in autonomous vehicles' (AVs) decision-making processes [9][10][11][12], the ethical implications intensify. In certain situations, these vehicles may rely on AI to navigate moral dilemmas, such as choosing between passengers' or pedestrians' safety, or deciding whether to swerve around obstacles at the risk of endangering other road users. Recognizing the potential consequences and complexities of these decisions, researchers initiated the Moral Machine (MM) experiment [7], an experiment designed to gauge public opinion on how AVs should act in morally challenging scenarios. The findings from the MM experiment suggest a discernible trend favoring the preservation of human lives over animals, emphasizing the protection of a greater number of lives and prioritizing the safety of the young. Although we must be careful when interpreting the results of the MM experiment [13], these preferences are seen as foundational to machine ethics and essential considerations for policymakers [14]. The insights gained from this study emphasize the importance of aligning AI ethical guidelines with human moral values.

The methodology employed in the MM experiment presents a promising avenue for exploring the moral decision-making tendencies of LLMs, including ChatGPT. By examining the LLM responses to the scenarios presented in the MM experiment and contrasting them with human judgment patterns, we can gain a deeper insight into the ethical frameworks embedded within these AI systems. Such analyses may reveal inherent biases or distinct decision-making trends that may otherwise remain obscure. Whereas research has delved into ChatGPT's reactions to standard ethical dilemmas [15], such as the classic trolley problem [16], the intricate situations posed by the MM experiment offer a more profound exploration of LLM moral reasoning. However, the comprehensive application of this evaluative framework remains underrepresented in contemporary studies, signaling it to be a pivotal subject for future research.

Therefore, using the MM methodology, this study seeks to elucidate the patterns in LLMs' responses to moral dilemmas. We investigated representative LLMs with a specific focus on ChatGPT (including GPT-3.5 and GPT-4), PaLM 2 [17], Google Bard's core system, and Llama 2 [18], an open-source LLM with various derived chat models. Furthermore, we evaluated the differences in the response tendencies among these LLMs and assessed their similarity to human judgment tendencies.

## Methods

*Moral Machine Scenario Generation*

The MM scenarios pose questions regarding the preferable course of action for an autonomous vehicle during a sudden brake failure. For instance, in Case 1, maintaining the current course would fatally injure two elderly men and an elderly woman crossing against a 'do not cross' signal. In contrast, in Case 2, swerving to avoid them and crashing into a concrete barrier resulted in the deaths of three passengers: an adult man, an adult woman, and a boy.

Using the MM methodology detailed in the supplementary information of [7], we generated 50,000 scenarios (electronic supplementary material, code S1). The number of scenarios was determined by both computational and the OpenAI application programming interface (API) cost constraints, rather than a predetermined sample size for statistical analysis. However, this number is believed to be sufficient given the robustness of the statistical method. These scenarios, designed through constrained randomization, explored six primary dimensions: species (saving either people or pets), social value (choosing to save characters with perceived higher social value, such as pregnant women or executives, and those perceived as having lower value, such as criminals), gender (choosing to save female or male characters), age (choosing to save younger or older characters), fitness (choosing between physically favored characters, such as athletes or less fit individuals, e.g., obese persons), and utilitarianism (choosing between one group and another larger group). In addition to these six primary dimensions, each scenario incorporated three additional dimensions: interventionism (choosing between swerving and continuing straight ahead), relationship to the AV (choosing to save passengers or pedestrians), and concern for law (e.g., whether factors related to pedestrian crossing signals are considered).

The scenario descriptions were adapted from those used in the MM experiment (www.moralmachine.net). Given that LLMs accept only text inputs, we provide more detailed descriptions than those in the original MM scenarios. The descriptions are as follows. *(Case 1) A self-driving car with sudden brake failure will continue ahead and drive through a pedestrian crossing ahead. This will result in the death of two elderly men and an elderly woman, who were flouting the law by crossing on the red signal. Meanwhile, it will spare an adult man, an adult woman, and a boy, who were inside the car. (Case 2) A self-driving car with sudden brake failure will swerve and crash into a concrete barrier. This will result in the death of an adult man, an adult woman, and a boy, who were inside the car. Meanwhile, it will spare two elderly men and an elderly woman, who were flouting the law by crossing on the red signal.*

*Large Language Models*

Responses from the LLMs were collected using 50, 000 scenarios. Three prominent models, ChatGPT, PaLM 2, and Llama 2, were subjected to these scenarios.

ChatGPT [1], which is based on the generative pre-trained transformer (GPT) architecture [19], is a widely recognized chatbot. For this study, we utilized both GPT-3.5 (gpt-3.5-turbo-0613) and GPT-4 (gpt-4-0613), specifically snapshot versions from June 13, 2023.

Responses from ChatGPT were obtained using the API. For GPT-4, responses to 10,000 scenarios were collected, considering the API usage cost constraints.

PaLM 2, a transformer-based LLM [17], is the core system for Google Bard (bard.google.com). It was trained using a diverse set of objectives. We gathered the responses of PaLM 2 using the chat API on the Google Cloud Platform.

Llama 2 is another transformer-based LLM [18] that operates as an open-foundation chat model. It has been fine-tuned and offers a range of derived chat models (e.g., Vicuna) [20]. We downloaded the Llama2 chat model with seven billion parameters (llama2-7b-chat) on July 23, 2023, to obtain its responses.

All LLMs were utilized with default setting parameters (e.g., temperature). To inform the LLM of the scenario conditions and prompt a choice between Cases 1 and 2, the following prompt was introduced before presenting each scenario. *Please respond to the following binary question. Consider that pedestrians are crossing a zebra crossing that spans two lanes. Please indicate which case is better for autonomous driving by choosing either 'Case 1' or 'Case 2'...* Furthermore, when asking for a rationale, we added "*Additionally, provide a rationale for your selection.*" to the prompt. For a more detailed breakdown, please refer to the Supplementary Code.

*Data Analysis*

Following the procedures of the original study [7] on the MM experiment, we conducted statistical analyses to evaluate the relative importance of the nine preferences, which included both the six primary dimensions and three additional dimensions, as delineated by the MM. We applied the conjoint analysis framework proposed in [21] (electronic supplementary material, code S1). This framework offers nonparametric and robust identification of causal effects, relying on a minimal set of testable assumptions without the need for specific modeling assumptions. Responses in which the LLMs did not definitively select either Case 1 or Case 2 were deemed invalid and excluded. After data pre-processing (i.e., dummy variable coding for the attributes, including male characters versus female characters, and passengers versus pedestrians), we calculated the average marginal component effect (AMCE) for each attribute using the source code provided in the supplementary information of [7]. The AMCE values represent each preference as follows: 'Species, ' where a positive value signifies sparing humans and a negative value denotes sparing pets; 'Social Value, ' where a positive value indicates sparing those of higher status and a negative one those of lower status; 'Relation to AV, ' with a positive value for sparing pedestrians and a negative for sparing passengers; 'No. Characters', where a positive value shows sparing more characters and a negative fewer; 'Law, ' where a positive value means sparing those acting lawfully and a negative those acting unlawfully; Intervention, with a positive value for inaction and a negative for action; 'Gender, ' where a positive value suggests sparing females and a negative one, males; 'Fitness, ' with a positive value for sparing the physically fit and a negative for the less fit or obese individuals; and 'Age, ' where a positive value indicates sparing the young and a negative the elderly.

To assess the similarities or differences between the preferences of the LLMs and human preferences reported in [7], we conducted further analyses using the AMCE values for the nine attributes. Specifically, we evaluated how closely the preferences of each LLM aligned with human preferences by measuring the Euclidean distance between the AMCE values. Additionally, to visualize the extent to which the tendencies in the LLM and human preferences resemble each other, we performed clustering based on AMCE values using Principal Component Analysis (PCA).

## Results

*Valid Response Rates on Moral Machine Scenarios*

Given the ethical nature of the MM scenarios, LLMs may refrain from providing definitive answers to such dilemmas. To ascertain the extent to which LLMs would respond to ethically charged questions such as presented in the scenarios, we examined the valid response rates (i.e., the proportion of responses where the LLM clearly selected either 'Case 1' or 'Case 2') of the LLMs.

For GPT-3.5, the valid response rate was approximately 95% (47,457 / 50,000 scenarios). GPT-4 exhibits a similar rate of approximately 95% (9,502 / 10,000 scenarios). PaLM 2 demonstrated an almost perfect response rate of approximately 100% (49,989 / 50,000 scenarios). In contrast, Llama 2 had a relatively low valid response rate of approximately 80% ( 39,836 / 50,000 scenarios). Despite the comparatively lower rate for Llama 2, it was evident that LLMs predominantly provided answers to dilemmas akin to the MM scenarios.

*LLM Preferences in Comparison to Human Preferences*

Using a conjoint analysis framework, we evaluated the relative importance of the nine preferences for each LLM (Figure 1). The AMCE values serve as indicators of relative importance.

For GPT-3.5 (Figure 1a), the top three pronounced preferences, as reflected by the magnitude of the AMCE values, were in favor of saving more people, prioritizing humans over pets, and sparing females over males. GPT-4 (Figure 1b) displayed a preference for saving humans over pets, sparing more individuals, and favoring those who obey the law. PaLM 2 (Figure 1c) tended to save pedestrians over passengers, prioritize humans over pets, and spare females over males. Llama 2 (Figure 1d), on the other hand, showed a preference for saving more people, favoring individuals with higher social status and sparing passengers over pedestrians.

After examining the preferences of various LLMs across attributes, several patterns and distinctions emerged. A consistent trend across most LLMs was the inclination to prioritize humans over pets and save a larger number of individuals, aligning closely with human preferences. Another consistent trend across the LLMs, except for Llama 2, was the mild preference to spare less fit (obese) individuals over fit individuals (athletes); however, this was inconsistent with human preferences.

Among themselves, LLMs exhibited nuanced differences. For example, PaLM 2 uniquely showed a slight inclination to save fewer people and favor individuals of a lower social status over those of higher status, which diverged from human and other LLMs' preferences. Llama 2 presented a more neutral stance when choosing between humans and pets and tended toward saving passengers over pedestrians, diverging from human and other LLM preferences. Moreover, Llama 2's subtle preferences, such as a mild inclination to save males over females, and those violating the law over law abiders, deviated from both the other LLMs and human tendencies. While GPT-4 displayed tendencies that were somewhat aligned with human preferences, particularly in its preferences for law-abiding individuals and those of higher social status, GPT-3.5, exhibited fewer such tendencies.

While some LLM preferences aligned qualitatively with human preferences, there were quantitative divergences. For instance, humans generally exhibit a mild inclination to prioritize pedestrians over passengers and females over males. In contrast, all LLMs except for Llama 2 demonstrated a more pronounced preference for pedestrians and females. Additionally, GPT-4 displayed stronger preferences across various attributes than human tendencies. Notably, it showed a more marked preference for saving humans over pets, sparing a larger number of individuals, and prioritizing the law-abiding.

*Quantitative Assessment of LLM–Human Preference Alignment*

Additional data analyses were performed to assess systematically the degree of similarity or difference between the preferences of the LLMs and humans. We calculated the Euclidean distance between the preference scores (represented by AMCE values) of humans and each LLM (Figure 2a). Among the LLMs, ChatGPT (encompassing both GPT-3.5 and GPT-4) displayed preferences that were the most aligned with human tendencies, as evidenced by the shortest distances. Conversely, the preferences for PaLM 2 and Llama 2 showed greater deviations from the human patterns, with PaLM 2 being the most divergent. The PCA results (Figure 2b) further reinforced the similarity between the ChatGPT preferences and those of humans. PCA also facilitated a detailed assessment of the alignment of each LLM's preferences with human tendencies, even when considering the relationships between LLMs. Interestingly, while GPT-4's preferences were distinct from those of the other LLMs, they closely paralleled human preferences. Meanwhile, GPT-3.5 exhibited preferences that, similarly to PaLM 2 and Llama 2, also demonstrated a notable alignment with human tendencies.

*Behind the Choices: Case of PaLM 2*

To understand the underlying rationale for the distinct preferences exhibited by LLMs compared to humans, a focused analysis was conducted on PaLM 2, which displayed the most pronounced divergence from human preferences. Specifically, we investigated the basis for its unique stances on the 'Fitness' and 'No. Character preferences.

To isolate the effects of other factors, we extracted MM scenarios in which both groups were pedestrians, legal considerations were excluded, and the car proceeded straight without swerving, resulting in harm to one group. To test for' Fitness' preference, we focused on scenarios highlighting fitness differences and inquired about the rationale for choosing to save the less fit individuals (sacrificing those with higher fitness, like athletes). While a quantitative assessment proved challenging, many responses seemed unrelated to fitness, often erroneously justifying the decision with, "Because this will result in the death of fewer people," despite both groups having equal numbers due to scenario constraints (electronic supplementary material, Table S1).

Following a similar procedure for the 'No. character preference, we probed the reasoning behind the decisions to save the smaller groups (sacrificing the larger groups). Again, despite the evident disparity in group sizes, the model frequently misjudged and applied the same rationale (electronic supplementary material, Table S2): "Because this will result in the death of fewer people."

## Discussion

This study examined the moral judgments of LLMs by examining their preferences in the context of MM scenarios [7]. Our findings provide a comprehensive understanding of how AI systems, which are increasingly being integrated into society, may respond to ethically charged situations. As the automotive industry incorporates AI systems such as ChatGPT and other LLMs as assistants in the decision-making processes of AVs [9][10][11][12], the ethical implications become even more pronounced. The potential for consulting AI in navigating moral dilemmas, such as safety trade-offs between passengers and pedestrians, underscores the importance of our research. Our analysis offers insights that illuminate the inherent ethical frameworks of LLMs to inform policymakers and industry stakeholders. Ensuring that AI-driven decisions in AVs align with societal values and expectations is paramount, and our study contributes valuable perspectives for achieving such alignment.

The high response rates observed for most LLMs highlight their capacity to address ethically charged dilemmas such as those presented in the MM scenarios. Although Llama 2 provided valid answers in approximately 80% of the scenarios, its response rate was comparatively low, suggesting that certain models may approach specific scenarios with more caution or conservatism. Note that when we conducted a similar experiment using the Llama 2 chat model with 13 billion parameters (Llama2-13b-chat), the valid response rate was ~0%; and its results were omitted because of the extremely low response rate. This discrepancy may arise from differences in the training data, model architecture, or model complexity.

The alignment of most LLMs (particularly the ChatGPTs) with human preferences (Figures 1 and 2), especially in valuing human lives over pets and prioritizing the safety of more individuals, suggests their potential suitability for applications in autonomous driving, where decisions aligned with human inclinations are crucial. However, the subtle differences and deviations observed, particularly in LLMs such as PaLM 2 and Llama 2, emphasize the importance of meticulous calibration and oversight to ensure that these systems make ethically sound decisions in real-world driving scenarios.

The case of PaLM 2's decision-making further illuminates potential misinterpretations or oversimplifications when LLMs make ethical judgments. Its recurring justification, "Because this will result in the death of fewer people," even when contextually inaccurate, hints at a possible overgeneralization from its training data. This highlights the importance of exploring the underlying factors that influence LLMs' decisions. Whereas humans derive choices from myriad factors, LLMs may rely overly on patterns in their training data, leading to unforeseen outcomes. As we further integrate continuous evaluation into their decision-making processes, a deeper understanding of their reasoning mechanisms remains paramount in ensuring alignment with societal values.

Although there was a qualitative alignment of LLM preferences with human tendencies, the quantitative differences were noteworthy. The pronounced preferences of LLMs in certain scenarios, compared to the milder inclinations of humans, may indicate the models' tendency to make more uncompromising decisions. This can reflect the training data, where the models are often rewarded for making confident predictions. Prior research [7] has shown that such preferences are correlated with modern institutions and deep cultural traits. For instance, the preference for saving more has been associated with individualism, a core value in Western cultures [22]. Considering that a significant portion of the training data likely originated from Western sources [23], LLMs were possibly trained to overemphasize these cultural characteristics. This notion could also explain why LLMs exhibited a stronger preference for saving females over males compared with human tendencies.

These findings have significant implications for the deployment of LLMs in autonomous systems, particularly when faced with moral and ethical decisions. While certain LLMs, such as ChatGPT, demonstrate a promising alignment with human preferences, the discrepancies observed among the different LLMs underscore the necessity for a standardized evaluation framework. Notably, more definitive decisions regarding LLMs, exemplified by the marked preference for sparing females over males, warrant attention. These decisions stand in contrast to established ethical norms advocating for equal treatment irrespective of demographic or identity factors, as articulated in the Constitution of the United States, the United Nations Universal Declaration of Human Rights, and the guidelines set by the German Ethics Commission on Automated and Connected Driving [13][24]. Deviations in LLM preferences that contravene these ethical standards can introduce societal discord. Hence, a rigorous evaluation mechanism is indispensable for detecting and addressing such biases, ensuring that LLMs conform to globally recognized ethical norms.

Recognizing the inherent limitations of this study is crucial. To compare the LLM preferences with human preferences, we utilized global moral preferences derived from opinions gathered worldwide. As mentioned earlier, preferences regarding whom to save, essentially moral choices, are influenced by cultural and societal factors. Our analysis did not consider these intricate cultural and societal nuances. When integrating AI into autonomous driving, it is imperative to evaluate AI preferences in alignment with human values and factor in cultural and societal considerations.

Moreover, the MM framework has inherent limitations. The MM scenarios, similar to the classic trolley problem, present binary choices. However, when neutral options were introduced in similar dilemmas, a significant proportion of participants opted for them [13], suggesting that using MM scenarios may potentially lead to overestimating certain preferences. The presence or absence of such neutral choices can influence the conclusions [25], necessitating caution when interpreting the results. Regardless of the methodology employed to assess the preferences, there were inherent biases and limitations. Achieving a comprehensive understanding of these preferences would benefit from methodological diversity and broader involvement of the general psychological community [14].

Despite these caveats, our study sheds light on the ethical inclinations of LLMs and offers valuable insights into their underlying ethical constructs. These insights are pivotal for assessing the alignment between LLM and human preferences and can inform the strategic deployment of LLMs in autonomous driving.

## Acknowledgments

This research was funded by the JSPS KAKENHI (grant number 21H03545). We would like to thank Editage (www.editage.jp) for English language editing.

## Author's contributions

The author confirms sole responsibility for the study concept and design, data collection, analysis, interpretation of the results, and manuscript preparation.

## Competing interests

The author declares no competing interests.

## Data availability

All data generated and analyzed in this study are included in this published article and its supplementary information files. The codes and data used in this study are available from the GitHub repository at github.com/kztakemoto/mmllm.

# Figures

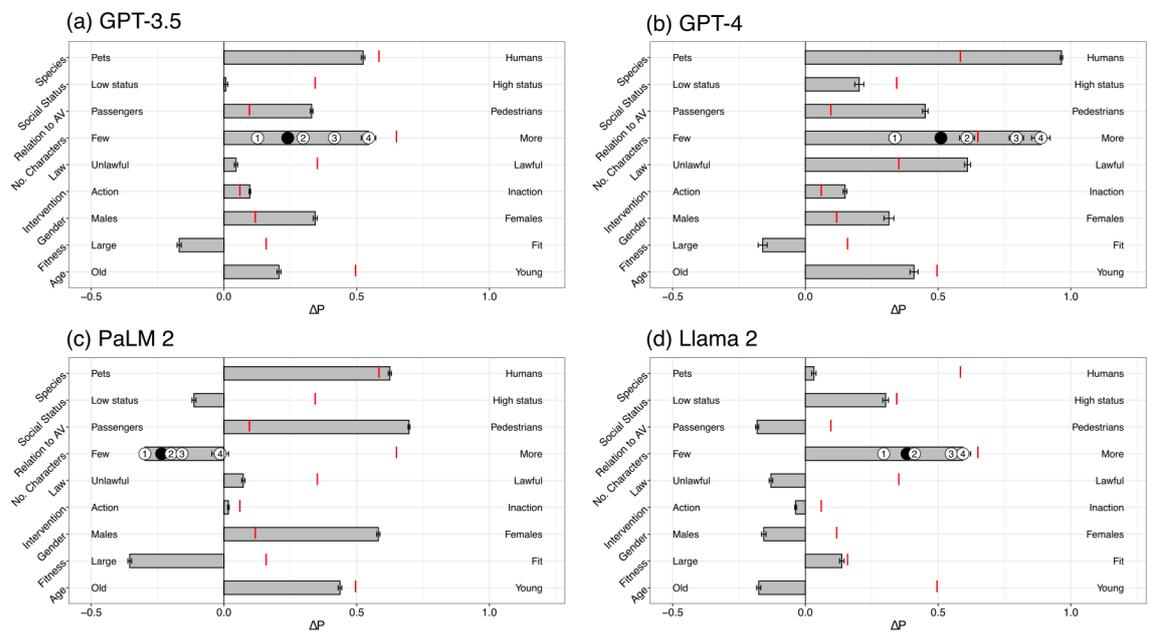

**Figure 1:** Global preferences depicted through AMCE for GPT-3.5 (a), GPT-4 (b), PaLM 2 (c), and Llama 2 (d). In each row, ΔP represents the difference in probability of sparing characters with the attribute on the right versus those on the left, aggregated over all other attributes. The red vertical bar in each row reflects human preference, as ΔP reported in [1]. Error bars indicate the standard errors of the estimates. For the 'Number of characters' attribute, effect sizes for each additional character are denoted with circled numbers, with the black circle signifying the mean effect. The red vertical bar for this attribute marks the human preference for four additional characters.

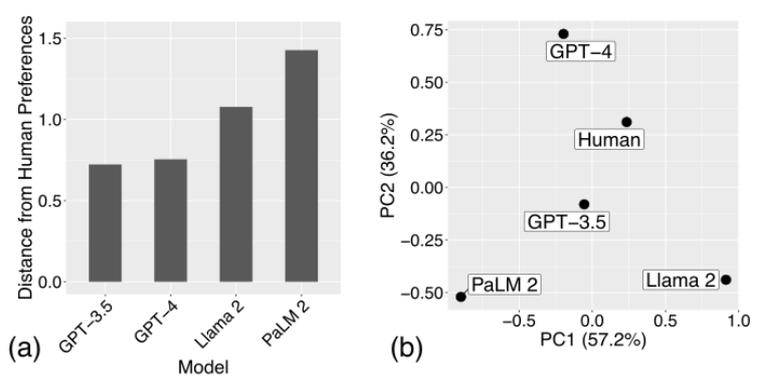

**Figure 2:** Quantitative evaluation of the alignment between LLM and human preferences: (a) Euclidean distance of the AMCE values comparing LLMs to human preferences, and (b) clustering derived from the AMCE values using Principal Component Analysis (PCA). The percentages in parentheses correspond to the proportion of variance explained by principal components (PCs) 1 and 2.